\def\year{2022}\relax
\documentclass[letterpaper]{article} % DO NOT CHANGE THIS
\usepackage{aaai22}  % DO NOT CHANGE THIS
\usepackage{times}  % DO NOT CHANGE THIS
\usepackage{helvet}  % DO NOT CHANGE THIS
\usepackage{courier}  % DO NOT CHANGE THIS
\usepackage[hyphens]{url}  % DO NOT CHANGE THIS
\usepackage{graphicx} % DO NOT CHANGE THIS
\usepackage{natbib}  % DO NOT CHANGE THIS AND DO NOT ADD ANY OPTIONS TO IT
\usepackage{caption} % DO NOT CHANGE THIS AND DO NOT ADD ANY OPTIONS TO IT
\DeclareCaptionStyle{ruled}{labelfont=normalfont,labelsep=colon,strut=off} % DO NOT CHANGE THIS
\usepackage{algorithm}
\usepackage{algorithmic}
\usepackage{amsmath}
\usepackage{subfigure}
\usepackage{multirow}
\usepackage{colortbl}

\usepackage{newfloat}
\usepackage{listings}
\floatstyle{ruled}
\newfloat{listing}{tb}{lst}{}
\floatname{listing}{Listing}
\title{Towards Efficient Data-Centric Robust Machine Learning with Noise-based Augmentation}
\author {
    Xiaogeng Liu,
    Haoyu Wang, 
    Yechao Zhang,
    Fangzhou Wu,
    Shengshan Hu
}
\affiliations {
    School of Cyber Science and Engineering\\
    National Engineering Research Center for Big Data Technology and System \\
    Services Computing Technology and System Lab\\
    Hubei Engineering Research Center on Big Data Security\\
    Huazhong University of Science and Technology\\
    Wuhan, 430074, China \\
    \{liuxiaogeng, wwwwhy, ycz, warkzhou, hushengshan\}@hust.edu.cn
    
} 
\begin{document}

\maketitle

\begin{abstract}
The data-centric machine learning aims to find effective ways to build appropriate datasets which can improve the performance of AI models. In this paper, we mainly focus on designing an efficient data-centric scheme to improve robustness for models towards unforeseen malicious inputs in the black-box test settings. Specifically, we introduce a noised-based data augmentation method which is composed of Gaussian Noise, Salt-and-Pepper noise, and the PGD adversarial perturbations. The proposed method is built on lightweight algorithms and proved highly effective based on comprehensive evaluations, showing good efficiency on computation cost and robustness enhancement. In addition, we share our insights about the data-centric robust machine learning gained from our experiments.
\end{abstract}

\section{Introduction}\label{intro}
With \textit{deep neural networks} (DNNs) being deployed in more and more fields, how to improve the performance of DNN models has attracted many researchers. A lot of efforts have been made to construct powerful models~\cite{DBLP:journals/corr/SimonyanZ14a,DBLP:conf/cvpr/HeZRS16,DBLP:conf/eccv/HeZRS16,DBLP:conf/bmvc/ZagoruykoK16} that can run accurately on different data from distinct tasks, and even outperform human's perception systems. However, the data, which can be regarded as “food” for the \textit{artificial intelligence} (AI) models, has not been sufficiently investigated as well as the model structures or the training functions. Recently, such status brings concerns from some academics and people start to research about the data and call it \textit{the data-centric machine learning}~\cite{miranda2021datacentric}. 

\begin{figure}[t]
\setlength{\belowcaptionskip}{-0.7cm}
\centering
\includegraphics[width=0.36\textwidth]{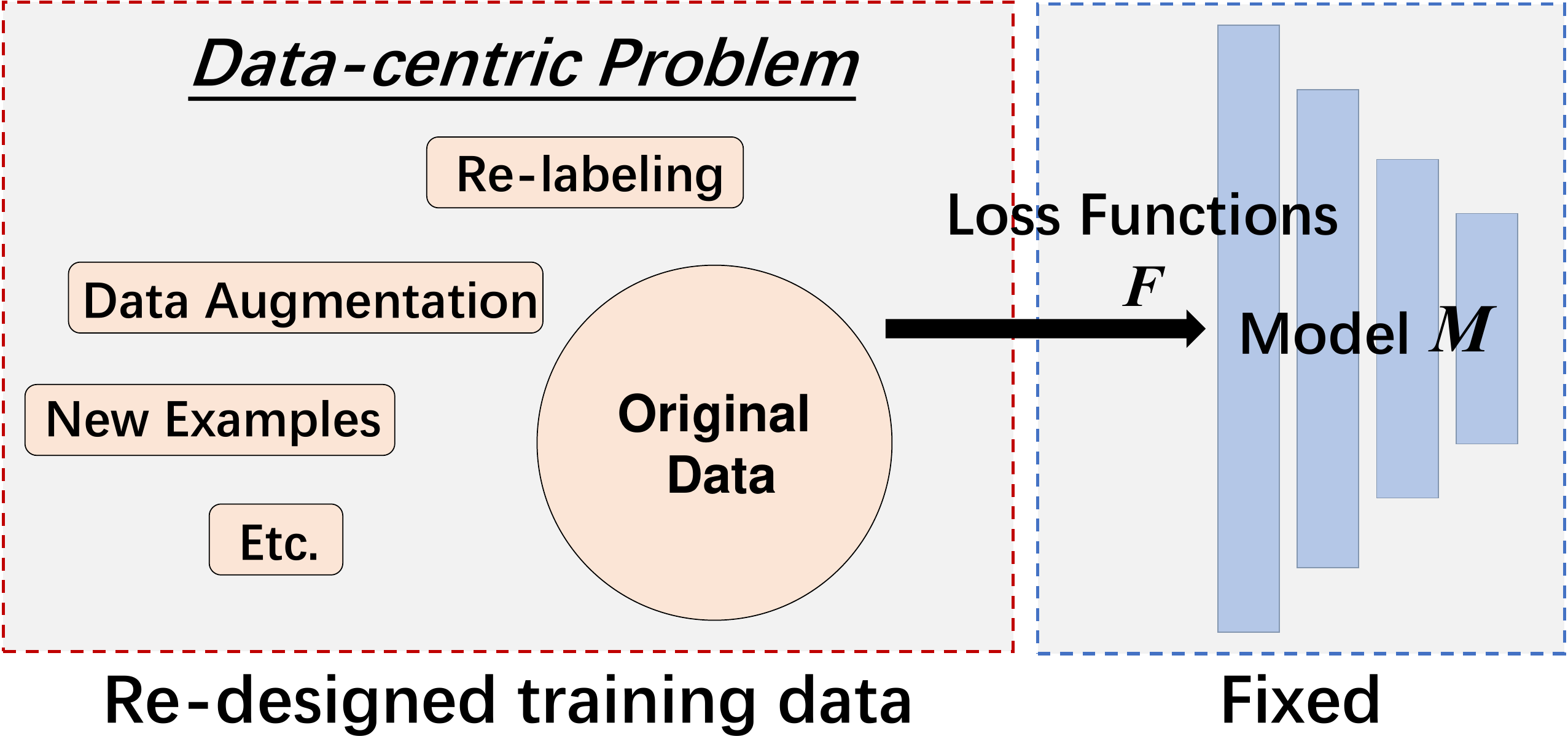}
\caption{Data-centric machine learning mainly focuses on the design of the dataset rather than model structure or loss function.}
\label{fig:introduction}
\end{figure}

As illustrated in Fig.~\ref{fig:introduction}, the data-centric machine learning is mainly focused on the design of data. The researchers are usually asked to construct elaborately designed datasets for the fixed model structure and loss functions and improve the performance of the models. The data-centric machine learning has many differences with the model-centric machine learning, turning researchers' eyes on data augmentation~\cite{DBLP:journals/jbd/ShortenK19} and feature engineering~\cite{dong2018feature}. 

In this paper, we mainly focus on how to design a dataset that can give robustness for the trained models towards any potential malicious inputs, i.e., the data-centric robust machine learning in the black-box settings. Besides, we want to keep our solution as simple as possible to achieve efficiency in practice. We first share our analysis about this topic as follows: in general, a data-centric problem can be solved by multiple data augmentation techniques, such as image noise, knowledge distillation with soft label~\cite{DBLP:journals/corr/HintonVD15}, augmentation by affine transformations. However, for knowledge distillation with soft label, its reliability has a dependence on the performance of the teacher model while we find it hard to choose (or train) an appropriate teacher for robust learning in the black-box settings. In addition, defensive distillation~\cite{DBLP:conf/sp/PapernotM0JS16} is proven to be vulnerable to specific adversarial attacks~\cite{DBLP:conf/sp/Carlini017}. For affine transformation, we argue that this method tends to empower the networks with robustness against angle transformation, partial information, etc., which are more likely to occur in the physical world. In this paper, we mainly focus on the digital world where image corruption and adversarial examples~\cite{DBLP:journals/corr/GoodfellowSS14,DBLP:conf/iclr/MadryMSTV18,DBLP:conf/mm/HuZLZLJ21,hu2022protecting} are more common and dangerous while affine transformations can hardly provide help on that. Consequently, we believe a more effective and efficient solution is using hard-label data with noise-based data augmentation including conventional image noise and adversarial perturbations~\cite{DBLP:journals/corr/GoodfellowSS14,DBLP:conf/iclr/MadryMSTV18}, forcing the model to learn robust features from corrupted and adversarial images.

 Another important and changeable component in data-centric problems is the settings of the (commonly-used) optimizer, especially for the learning rate scheduler. In our method, we choose the cosine annealing schedule with warm restart~\cite{DBLP:conf/iclr/LoshchilovH17}. Compared with other schemes and vanilla cosine annealing, this scheduler scans a wider range of learning rate values, provides a better generalization and accuracy, and accelerates the training process of models. In 
a nutshell, our solution is composed of noise-based data augmentation and cosine annealing learning rate scheduler with warm restart.

We summarize our contributions as follows:
\begin{itemize}
\item We introduce a noise-based method to construct a comprehensive training dataset to empower the trained model with strong robustness without dependence on any additional training process or loss function. In addition, this scheme is composed of lightweight algorithms and achieves efficiency in terms of effectiveness and computation cost.
\item Specifically, we evaluate the performance of different image noise and adversarial perturbations for data augmentation comprehensively. Based on the concrete quantitative results, we choose Gaussian Noise, Salt-and-Pepper noise, and the PGD perturbations as integrated solutions.
\item Our method shows good effectiveness on evaluations and ranks $8$-th place out of $3,691$ teams on  \textit{AAAI2022 Security AI Challenger\footnote{https://tianchi.aliyun.com/competition/entrance/531939/information}: Data Centric Robust Learning on ML models}, surpassing the baseline by $20.03\%$ in terms of final score.
\end{itemize}

\section{Methodology}\label{method}

\subsection{Noise-based Data Augmentation}\label{Analysis}
Here we highlight four kinds of image noise for data augmentation, including two kinds of conventional image noise (Salt$\And$Pepper, Gaussian) and two kinds of adversarial perturbations (FGSM, PGD). 

\textbf{Salt and Pepper noise. }In a corruption image, the Salt\&Pepper noise is presented as pure white or black pixel with discrete distribution, which is often caused by sharp and sudden disturbances in the image signal. We argue that it also serves as random masks and makes models learn partial features for robustness.

\textbf{Gaussian Noise. }Adding Gaussian noise is a widely-used method for data augmentation. The probability density function of Gaussian noise follows the normal distribution (i.e, Gaussian distribution).

\textbf{FGSM. }The Fast Gradient Sign Method (FGSM)~\cite{DBLP:journals/corr/GoodfellowSS14} is a single-step gradient-based adversarial attacks scheme, which calculates the gradient sign of output loss from a white-box model and then computes (usually $l_{\infty}$-norm) bounded adversarial perturbations as we demonstrate as follows:
\begin{equation}\label{eq:FGSM}
\begin{aligned}
    x_{adv} = x + \eta,\medspace \eta=\epsilon \operatorname{sign}\left(\nabla_{x} J(\theta, x, y)\right),
\end{aligned}
\end{equation}
where $J$ represents the loss function of the white-box model, $\epsilon$ is the perturbation size, and $y$ is the ground-truth label of the clean image $x$.

\textbf{PGD. }The Projected Gradient Descent (PGD)~\cite{DBLP:conf/iclr/MadryMSTV18} is a strong iterative adversarial attack which generates adversarial examples by:
\begin{equation}\label{eq:PGD}
\begin{aligned}
    x^{t+1}=\Pi_{x+\mathcal{S}}\left(x^{t}+\alpha \operatorname{sign}\left(\nabla_{x} J(\theta, x, y)\right)\right),
\end{aligned}
\end{equation}
where $\alpha$ is the perturbation size per-step, and $\Pi_{x+\mathcal{S}}$ represents the similarity restriction which is usually a clip function in practice.

It can be noted that both FGSM and PGD are calculated on a white-box model, which means a trained model is needed when we construct the new data from the original ones. In our method, we first train the networks on the original clean dataset, then we generate adversarial examples (FGSM, PGD) on them. This leads to a little difference from recent adversarial training~\cite{DBLP:conf/iclr/PangYDSZ21} where adversarial examples are calculated in the training process simultaneously, while our method needs to train a clean model in the first stage due to the specific requirements of data-centric scenario. 

\begin{figure}[t]
\setlength{\belowcaptionskip}{-0.5cm}
\centering
\includegraphics[width=0.36\textwidth]{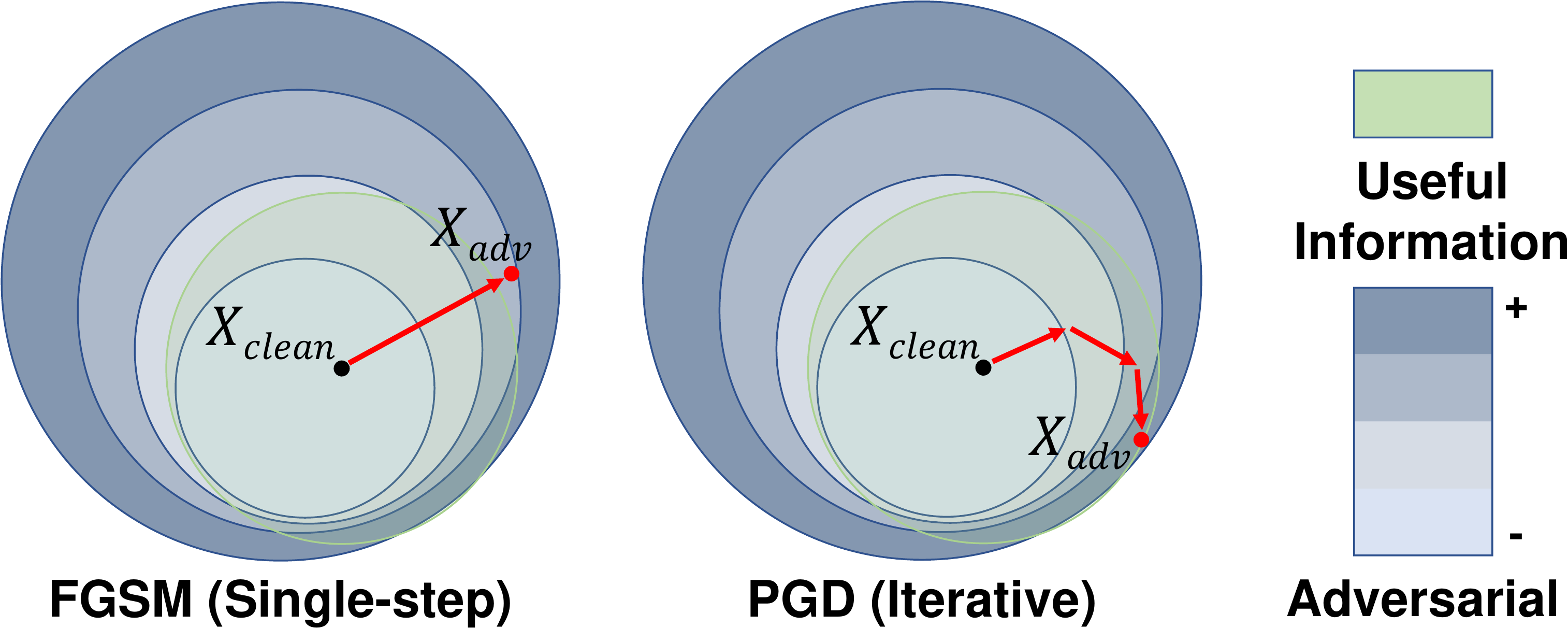}
\caption{Illustration of “adversarial under-fitting” phenomenon. The darker blue represents the stronger attack ability of an adversarial example. The green region represents a range where the adversarial example can provide comprehensively useful information as training data. Due to the single-step design, the example generated by FGSM is likely to locate on a point with weaker adversarial strength, meanwhile losses more clean features.}
\label{fig:method}
\end{figure}

Although we cannot modify training process like adversarial training, we argue that the choice of adversarial algorithms in data-centric machine learning is the same as what is widely adopted in adversarial training. In adversarial training, a majority of researchers use PGD~\cite{DBLP:conf/iclr/PangYDSZ21,DBLP:conf/icml/ZhangYJXGJ19,DBLP:conf/icml/RiceWK20} to generate adversarial examples, since PGD is proven to have stronger attack strength and unlikely to cause label leaking~\cite{DBLP:conf/iclr/MadryMSTV18}. Due to the single-step design, the attack strength of FGSM is usually weaker than PGD in the same perturbation size, and it also tends to create adversarial examples with more conspicuous changes. This means that when it comes to training models straightly on adversarial examples, FGSM may drive the model far from an optimal point where the model can handle different inputs including clean, corrupted, and adversarial images. As Fig.~\ref{fig:method} shows, compared with FGSM, PGD is more likely to have stronger adversarial strength and maintain more clean features. We leave a more comprehensive investigation about this in the experiments.

\begin{algorithm}[h]
\caption{SGD with CosineAnnealingWarmRestarts}
\label{alg}
\textbf{Input}: Parameters $\theta_t$; minimum learning rate $\gamma_{min}$; maximum learning rate $\gamma_{max}$; loss function $\Phi$; weight decay $\lambda$; momentum $\mu$; dampening $\tau$; the number of epochs since the last restart $T_{cur}$; the number of epochs between two warm restarts $T$.\\
\textbf{Output}: Parameters $\theta_{t+1}$.
\begin{algorithmic}[1] 
\STATE \textit{CosineAnnealingWarmRestarts Schedule.}
\IF{$T_{cur} == T$}
\STATE $\gamma = \gamma_{min}$;
\ELSIF{$T_{cur} == 0$}
\STATE $\gamma = \gamma_{max}$;
\ELSE
\STATE $\gamma = \gamma_{min} + \frac{1}{2}(\gamma_{max} - \gamma_{min})(1 + \cos{(\frac{T_{cur}}{T}\pi)})$;
\ENDIF
\STATE \textit{Stochastic Gradient Descent.}
\STATE $g_{t+1} \gets \nabla_{\theta}\Phi(\theta_{t})$
\STATE $g_{t+1} \gets g_{t+1} + \lambda\theta_{t}$
\STATE $g_{t+1} \gets \mu g_{t} + (1 - \tau)g_{t+1}$
\STATE $\theta_{t+1} \gets \theta_{t} - \gamma g_{t+1}$
\RETURN $\theta_{t+1}$
\end{algorithmic}
\end{algorithm}

\begin{figure*}[t]
\setlength{\belowcaptionskip}{-0.5cm}
\centering
\includegraphics[width=0.82\textwidth]{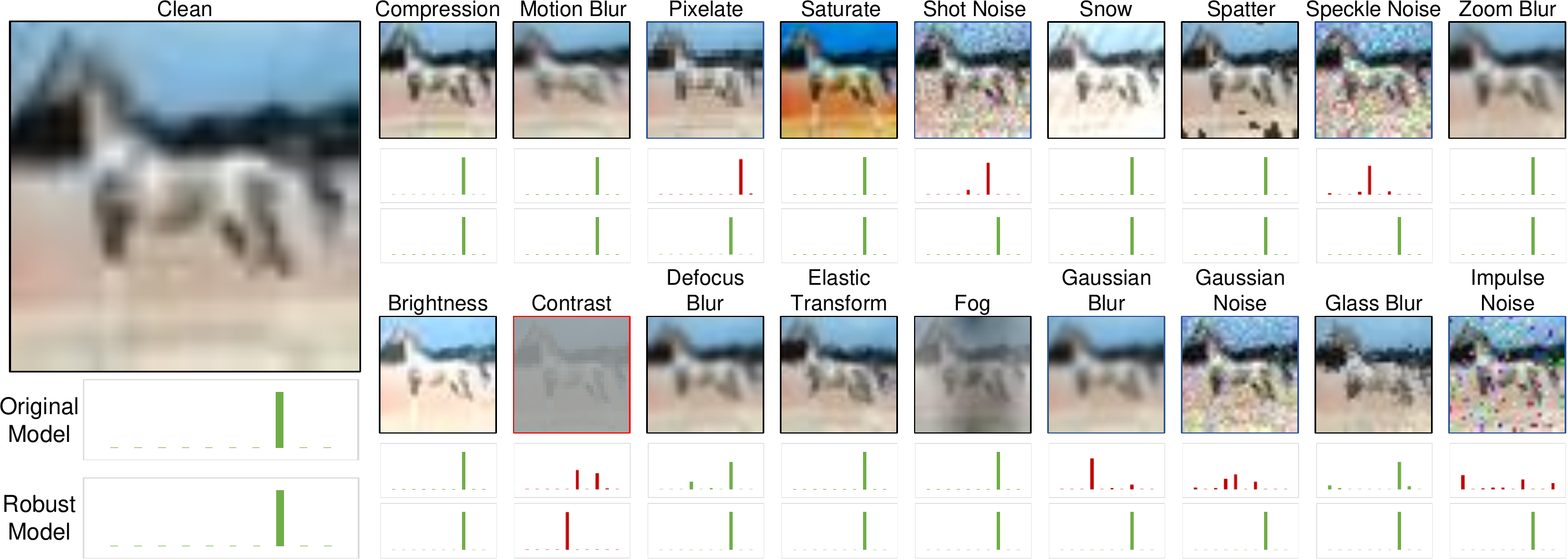}
\caption{Images from the official test data of \textit{AAAI2022 Security AI Challenger}. The bar charts below each image represent the softmax outputs from the original model (trained on the clean dataset) and the robust model (trained on our augmented dataset). The incorrect results are colored red. The softmax outputs show that our method enhances the robustness of the networks. And the enhanced model can give correct results with high confidence.}
\label{fig.visualize}
\end{figure*}
\subsection{Cosine Annealing and Warm Restart}
Cosine annealing combined with warm restart~\cite{DBLP:conf/iclr/LoshchilovH17} is a popular and effective technique to solve optimization problems which accelerates the convergence and avoids ill-condition problems. Specifically, as demonstrated in Alg.~\ref{alg}, this scheme sets the learning rate of the optimizer according to the cosine annealing curve and restart after some iterations. In our method, we leverage this technique to get a powerful Stochastic Gradient Descent (SGD) optimizer.

\section{Experiments}\label{experiment}
\subsection{Experimental Setting}
\textbf{Implementation details. }In our experiments, we construct the training data with four noise-based augmentation methods, i.e., Gaussian noise, Salt\&Pepper noise, FGSM~\cite{DBLP:journals/corr/GoodfellowSS14} perturbations, and PGD~\cite{DBLP:conf/iclr/MadryMSTV18} perturbations. For Gaussian noise, we set mean equals $0$ and variance equals $0.005$. For Salt\&Pepper noise, we set the noise ratio to $40\%$. For FGSM, we set perturbation size to $8/255$, $12/255$, $16/255$ respectively for the comparison study. For PGD, we choose the maximum perturbation equals $8/255$, $12/255$, $16/255$, epochs equal to $20$, and step size equals $0.4/255$, $0.6/255$, $0.8/255$ respectively. All these augmentations will be conducted upon the CIFAR10~\cite{krizhevsky2009learning} training data, which contains 50,000 $32\times32$ color images from 10 categories.

\textbf{Models. }Following the semi-finals of \textit{AAAI2022 Security AI Challenger}\footnote{https://tianchi.aliyun.com/competition/entrance/531939/information} competition, we use PreActResnet18~\cite{DBLP:conf/eccv/HeZRS16} and WideResnet~\cite{DBLP:conf/bmvc/ZagoruykoK16} as the test model.

\textbf{Test dataset. }We choose the official test dataset of \textit{AAAI2022 Security AI Challenger} competition, which is a mixed dataset containing clean images, corruption images, and adversarial images. All of these images are extracted/generated from the original CIFAR10 test set. The corruption image set contains 18 types of corruption, and the adversarial image set is generated from 3 different settings by model-ensemble~\cite{DBLP:conf/cvpr/DongLPS0HL18}. Visualized in Fig.~\ref{fig.visualize}.

\textbf{Optimizer and scheduler. }In our experiments, we use a SGD~\cite{Robbins2007ASA} optimizer with the \textit{CosineAnnealingWarmRestarts}\footnote{https://pytorch.org/docs/master/generated/torch.optim.lr\_scheduler.C\\osineAnnealingWarmRestarts.html} scheduler to train the test model. Specifically, for SGD, we set the learning rate equals $0.01$, momentum equals $0.9$, and weight decay equals $5\times10^{-4}$. For the scheduler, we set $T_0 = 3$, $T_{mult} = 2$, and the minimum learning rate equals $1\times10^{-5}$. In addition, we set the batch size equals $128$ and training epochs equal $185$.

\subsection{Comparison Study}

\begin{table*}[t]
\setlength{\belowcaptionskip}{-0.6cm}
\centering
\resizebox{0.95\textwidth}{!}
{
\begin{tabular}{c|c|ccccccccc}
    \hline 
    Test Data & Model & CL & GS & S\&P & FGSM$_{8}$ & FGSM$_{12}$ & FGSM$_{16}$ & PGD$_{8}$ & PGD$_{12}$ & PGD$_{16}$ \\
    \hline
    Clean  & PreActResnet18 & \cellcolor[rgb]{.682,  .667,  .667} 0.9922  & \cellcolor[rgb]{ .916,  .912,  .912} 0.9391  & \cellcolor[rgb]{ .851,  .851,  .851} 0.9643 & 0.3888  & 0.2993  & 0.1869  & 0.7302  & 0.6121  & 0.4579  \\
    Images & WideResnet & \cellcolor[rgb]{.682,  .667,  .667} 0.9923  & \cellcolor[rgb]{ .916,  .912,  .912} 0.9350  & \cellcolor[rgb]{ .851,  .851,  .851} 0.9751 & 0.4518  & 0.2772  & 0.2388  & 0.8776  & 0.7515  & 0.5910  \\
    % \hline
    Corrupted & PreActResnet18 & \cellcolor[rgb]{ .916,  .912,  .912} 0.6353  & \cellcolor[rgb]{ .851,  .851,  .851} 0.6971  & \cellcolor[rgb]{.682,  .667,  .667} 0.7162 & 0.2354  & 0.1981  & 0.1419  & 0.4410  & 0.4180  & 0.3818  \\
    Images & WideResnet & \cellcolor[rgb]{ .916,  .912,  .912} 0.6108  & \cellcolor[rgb]{ .851,  .851,  .851} 0.6982  & \cellcolor[rgb]{.682,  .667,  .667} 0.7265 & 0.3124  & 0.2143  & 0.1638  & 0.5986  & 0.4961  & 0.4315  \\
    % \hline
    Adversarial & PreActResnet18 & 0.3719  & 0.6433  & 0.5411 & 0.2758  & 0.2149  & 0.1591  & \cellcolor[rgb]{.682,  .667,  .667} 0.7860  & \cellcolor[rgb]{ .851,  .851,  .851} 0.7500  & \cellcolor[rgb]{ .916,  .912,  .912} 0.6922  \\
    Images & WideResnet & 0.3797  & 0.6506  & 0.5329 & 0.3455  & 0.2387  & 0.1984  & \cellcolor[rgb]{.682,  .667,  .667} 0.8052  & \cellcolor[rgb]{ .851,  .851,  .851} 0.7754  & \cellcolor[rgb]{ .916,  .912,  .912} 0.7211  \\
    \hline
    Avg.   & Final Score & 0.6637  & \cellcolor[rgb]{.682,  .667,  .667} 0.7606  & \cellcolor[rgb]{ .851,  .851,  .851} 0.7427 & 0.3350  & 0.2404  & 0.1815  & \cellcolor[rgb]{ .916,  .912,  .912} 0.7064  & 0.6338  & 0.5459  \\
    \hline
\end{tabular}%
}
\caption{Comparison of noise-based data augmentation methods. Each column represents the training data ($50,000$ images) augmented by a single method. We highlight the top-3 for each kind of test data and the final score. Notations CL: clean images. GS: images with Gaussian noise. S\&P: images with Salt\&Pepper noise. The subscript number ($8, 12, 16$) represents the perturbation size with $8/255, 12/255, 16/255$ for FGSM and PGD, respectively. We will use the same notations hereafter.}
\label{tab:comparison_noise}%
\end{table*}%

\textbf{Evaluations of different methods. }Tab.~\ref{tab:comparison_noise} shows the comparison study of different noise-based augmentation methods. In this experiment, we evaluate the model trained on the dataset which contains $50,000$ images constructed from the same original data with a single augmentation method, e.g., the GS represents the training dataset containing $50,000$ images only with Gaussian Noise and the FGSM$_{12}$ represents the training dataset containing $50,000$ images only with FGSM perturbations with perturbation size  = $12/255$. As shown in Tab.~\ref{tab:comparison_noise}, Gaussian noise, Salt\&Pepper noise, and the PGD perturbations are appropriate methods for robustness enhancement in terms of the final score. Salt\&Pepper noise shows superior effectiveness for corrupted images and the PGD perturbations have outstanding performance on the robustness towards adversarial images. Gaussian noise has a relatively balanced accuracy for each kind of image set, leading to its triumph on the final score.

Another non-negligible fact is that FGSM is not competitive to PGD for data-centric robust machine learning in the same perturbation size and this verifies our analysis in Fig~\ref{fig:method}. We also argue that FGSM in smaller perturbation sizes may be more appropriate for data augmentation.

\textbf{Evaluations of mixed dataset. }According to the performance of different noise-based augmentation methods in Tab.~\ref{tab:comparison_noise}, we construct 10 mixed datasets with different proportions of the noise methods as shown in Tab.~\ref{tab:notations}:

\begin{table}[h]
\setlength{\belowcaptionskip}{-0.3cm}
\centering
\resizebox{0.48\textwidth}{!}
{
\begin{tabular}{r|c|c|c|c|c|c|c|c|c|c}
    \hline 
     \textit{Mix.} & 1 & 2 & 3 & 4 & 5 & 6 & 7 & 8 & 9 & 10\\
    \hline
    Clean  & - & - & - & - & - & - & - & 10k & - & 10k\\
    GS & 10k & 10k & 10k & 10k & 10k & 10k & 10k & 10k & 10k & 10k\\
    S\&P & 10k & 10k & 10k & 10k & 10k & 10k & 10k & 10k & 10k & 10k\\
    FGSM$_{8}$  & 30k & - & - & - & - & - & 10k & 10k & - & -\\
    FGSM$_{12}$  & - & 30k & - & - & - & - & 10k & 10k & - & -\\
    FGSM$_{16}$  & - & - & 30k & - & - & - & 10k & - & - & -\\
    PGD$_{8}$  & - & - & - & 30k & - & - & - & - & 10k & 10k\\
    PGD$_{12}$  & - & - & - & - & 30k & - & - & - & 10k & 10k\\
    PGD$_{16}$  & - & - & - & - & - & 30k & - & - & 10k & -\\
    \hline
\end{tabular}%
}
\caption{Contents in 10 different mixed datasets.}
\label{tab:notations}%
\end{table}%

As shown in Fig.~\ref{Fig.mixed_scores}, an obvious tendency is that the datasets mixed with PGD perturbations, Gaussian noise, and Salt\&Pepper noise will reach a balance between robustness and accuracy as they all have relatively high final scores, among which the dataset with PGD$_{8/255}$ achieves the best performance.

\subsection{Ablation Study}
\begin{figure}[h] 
\setlength{\belowcaptionskip}{-0.6cm}
\centering
\subfigure[Final scores of different mixed dataset]{
\label{Fig.mixed_scores}
\includegraphics[width=0.45\textwidth]{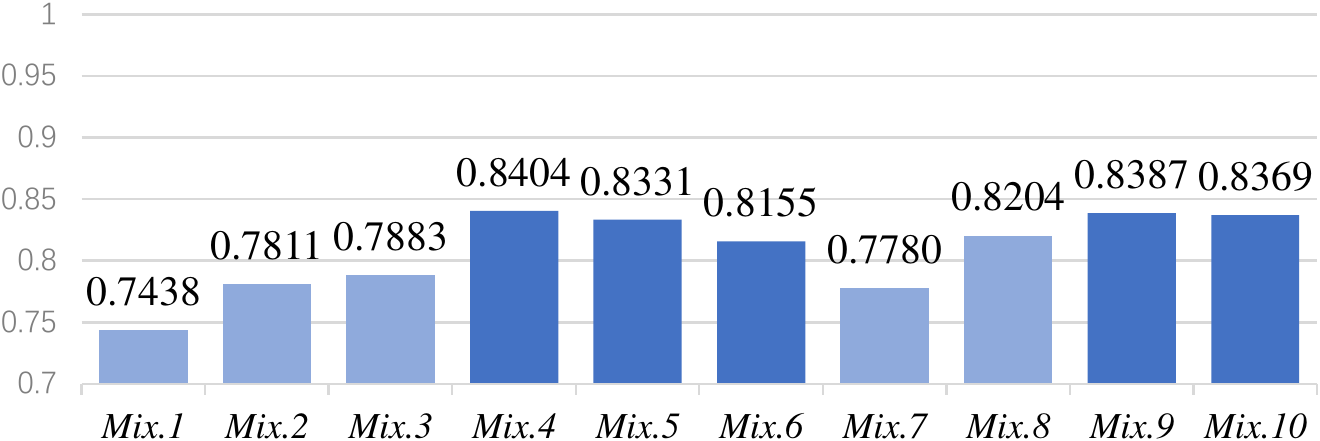} 
}
\subfigure[Ablation study of warm restarts]{
\label{Fig.ablation_scores}
\includegraphics[width=0.45\textwidth]{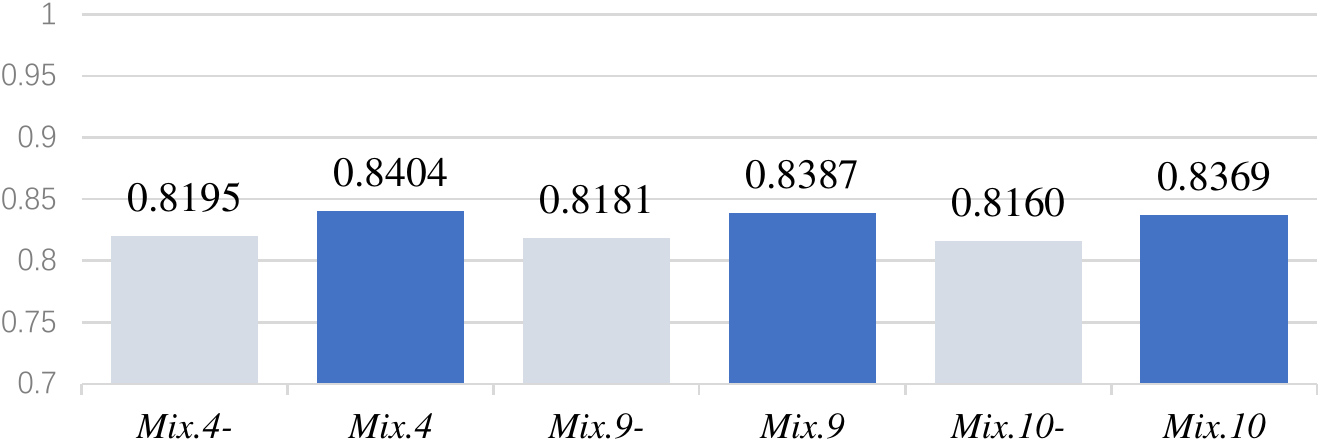} 
}
\caption{Notations are introduced in Tab.~\ref{tab:notations}. (a) shows the final scores of different mixed datasets, the datasets with PGD are colored darker blue. (b) represents the final scores of the top-3 datasets in Fig.~\ref{Fig.mixed_scores} with and without (bars with light grey) warm restarts design (but all have cosine annealing), showing that warm restarts leads to about $2\%\sim3\%$ improvement.}
\label{Fig.chart_bar}
\end{figure}

Fig.~\ref{Fig.ablation_scores} shows the ablation study of the warm restarts design. In this experiment, we choose three mixed datasets which have the highest final scores in our evaluations and train the test model by cosine annealing scheduler without warm restarts. As illustrated in Fig.~\ref{Fig.ablation_scores}, the warm restarts design can improve the model performance by $2\%\sim3\%$, which is a significant progress. 

\section{Conclusions}\label{conclusion}
\begin{itemize}
\item An appropriate scheme to generate adversarial examples is crucial. We argue that the samples generated by iterative algorithms (i.e., PGD) with a relatively small size of perturbations can maintain considerable clean features and avoid insufficiency of adversarial strength. 
\item Noise-based augmentation can perform better in data-centric settings compared with soft labels or affine transformations. And the appropriate noise type and hyper-parameters can ensure that the model is valid for both clean and perturbed images.
\item \textit{CosineAnnealingWarmRestarts} scheduler can avoid overfitting of the neural networks on limited data by restarting the learning rate, and thus enhance the generalization of the model. This method will lead to considerable improvements when it is combined with noise-based data augmentation.
\end{itemize}

% Use \bibliography{yourbibfile} instead or the References section will not appear in your paper
\clearpage
\bibliography{aaai22.bib}
\end{document}